%% file: main.tex
\documentclass[10pt,twocolumn,letterpaper]{article}

\usepackage{iccv}              

\usepackage{graphicx}
\usepackage{amsmath}
\usepackage{amssymb}
\usepackage{bm}
\usepackage{booktabs}
\usepackage{comment}
\usepackage{xcolor}
\usepackage{siunitx}
\graphicspath{{FiguresOverleaf/}}
\def\capfont{\normalfont\small}

\input{preamble}

\input{ALL_FIGURES}
\input{ALL_TABLES}

\input{sec/math_macros}

\definecolor{iccvblue}{rgb}{0.21,0.49,0.74}
\usepackage[pagebackref,breaklinks,colorlinks,allcolors=iccvblue]{hyperref}

\def\paperID{5786} 
\def\confName{ICCV}
\def\confYear{2025}

\title{Learning to See Inside Opaque Liquid Containers using Speckle Vibrometry}

\author{Matan Kichler \qquad Shai Bagon \qquad Mark Sheinin\\
Weizmann Institute of Science, Israel\\
{\tt\small matankic@gmail.com, shai.bagon@weizmann.ac.il, mark.sheinin@weizmann.ac.il}\\
{\small
\url{https://matankic.github.io/see_inside_containers}}
}

\begin{document}
\maketitle

\input{sec/abstract}

\section{Introduction}
\label{sec:intro}
\input{sec/intro}


\section{Background}
\label{sec:back}
\input{sec/back}

\section{Sensing rapid vibrations on a 2D grid}
\label{sec:camera}
\input{sec/cam}

\section{Learning to infer a container's liquid level}
\label{sec:learning}
\input{sec/learning}

\section{Implementation details}
\label{sec:details}
\input{sec/details}

\section{Experimental evaluation}
\label{sec:experiments}
\input{sec/experiments}

\section{Discussion and limitations}
\label{sec:discussion}
\input{sec/discussion}

\section{Conclusion}
\label{sec:conclude}
\input{sec/conclude}

\iftoggle{iccvfinal}{
\paragraph{Acknowledgments}
We thank G.~Shasha for experimental support.  
This work was supported by the Weizmann Center for New Scientists, the Institute of AI, and the MBZUAI-WIS Joint Program (SB).
}
\newpage
{
    \small
    \bibliographystyle{ieeenat_fullname}
    \bibliography{main}
}

\end{document}

%% file: preamble.tex
\usepackage{newtxmath}
\usepackage[scr=rsfso]{mathalfa}
\usepackage{multirow}
\usepackage{xspace}
\usepackage{ifthen}
\usepackage[normalem]{ulem}

\newcommand{\SB}[2][]{%
    \ifthenelse{ \equal{#1}{} }
        {\textcolor{orange}{#2\xspace{}}}
        {\textcolor{orange}{\sout{#1} #2\xspace{}}}
}
\newcommand{\MT}[2][]{%
    \ifthenelse{ \equal{#1}{} }
        {\textcolor{blue}{#2\xspace{}}}
        {\textcolor{blue}{\sout{#1} #2\xspace{}}}
}
\newcommand{\MS}[2][]{%
    \ifthenelse{ \equal{#1}{} }
        {\textcolor{red}{#2\xspace{}}}
        {\textcolor{red}{\sout{#1} #2\xspace{}}}
}

\def\eg{\emph{e.g.}} 

\def\ie{\emph{i.e.}} 

\def\etal{\emph{et al.}}

\newcommand{\afterfigure}{\vspace*{-2mm}}
\newcommand{\parvspace}{\vspace*{-12pt}}

\newcommand{\new}[1]{{\color{black}#1}}


%% file: ALL_FIGURES.tex
\newcommand{\figTeaser}{
\begin{figure}[t]
    \centering
    \includegraphics[width=\columnwidth]{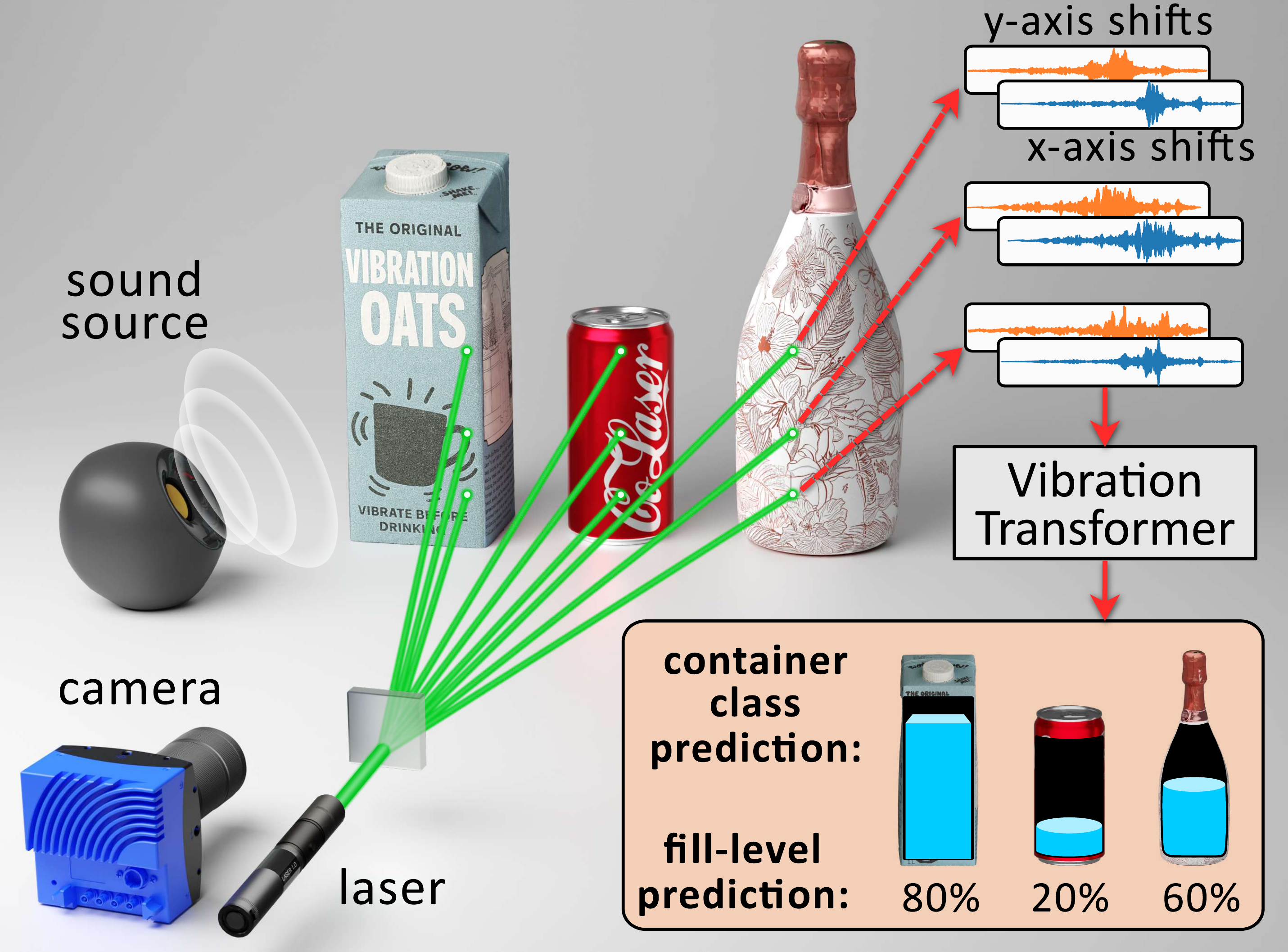}
    \caption{Learning to `see' the fill level of opaque containers. We develop a novel system to capture the surface vibrations of multiple liquid containers at multiple surface points. We vibrate the containers using a nearby speaker and train a novel Vibration Transformer to infer the container class and the hidden liquid level, expressed as a percentage of total capacity, from the measured multi-point surface vibrations. 
    }
    \afterfigure
    \label{fig:opening}
\end{figure}
}

\newcommand{\figWine}{
\begin{figure}[t]
    \centering
    \includegraphics[width=0.95\columnwidth]{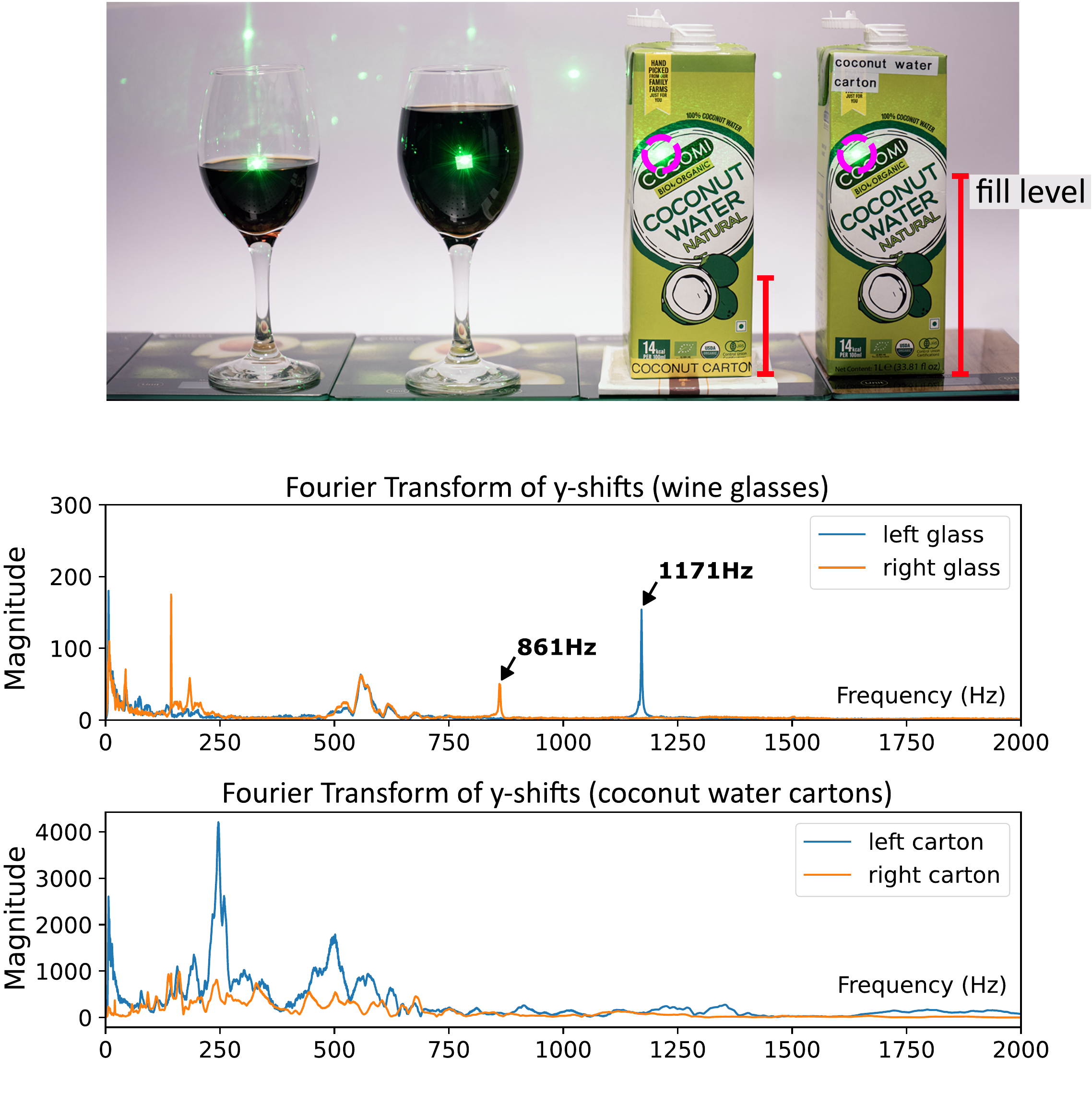}
    \put(-160,139.0){\capfont \textbf{(a)} single point experiment}
    \put(-175,1.0){\capfont \textbf{(b)} Fourier transforms of y-axis shifts}
    \caption{ 
    Variations in container responses.  
    \textbf{(a)} Setup showing vibration measurements from a resonant wine glass (clear for illustration) and a standard beverage container. We record responses from two identical items at different fill levels using a single surface point per object.  
    \textbf{(b)} The wine glass exhibits a distinct resonance that shifts with fill level. In contrast, the coconut water container exhibits a complex frequency response that is difficult to relate to its fill level. We train a transformer to infer fill levels from such signals by analyzing vibrations at \textit{multiple} surface points.} 
    \afterfigure
    \label{fig:wine}
\end{figure}
}

\newcommand{\figSystem}{
\begin{figure}[t]
    \centering
    \includegraphics[width=\columnwidth]{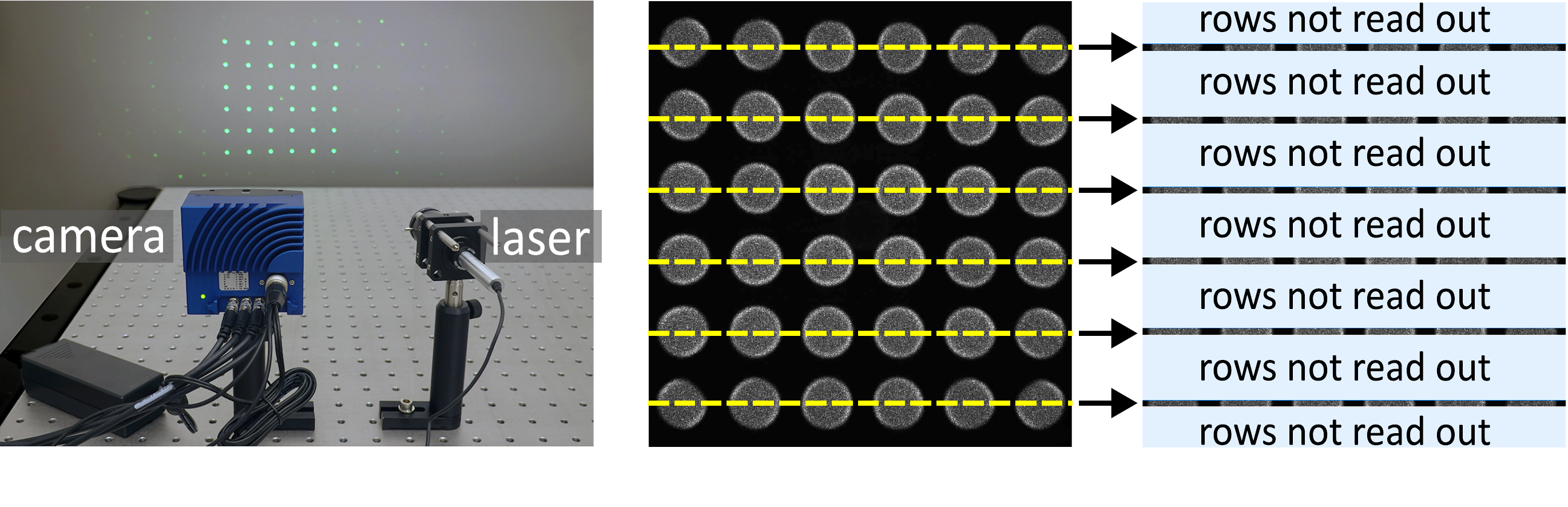}
    \put(-235,0.0){\capfont \textbf{(a)} simplified prototype}
    \put(-140,0.0){\capfont \textbf{(b)} full resolution}
    \put(-65,0.0){\capfont \textbf{(c)} defined ROIs}
    \caption{Capturing vibrations on a 2D grid. \textbf{(a)} Our system consists of a laser projecting a 2D grid of points on the scene, and a single defocused camera. The camera image, shown in (b), captures the speckle interference for all grid points. \textbf{(b)} We define a set of regions of interest (ROIs) centered on the middle of each defocused row of laser points (ROI centers marked in yellow), with each ROI having a height of a few pixels (\eg, six). 
    \textbf{(c)} The camera outputs only the concatenated ROIs, enabling high sampling rates of speckle vibrations (e.g., \SI{57}{\kilo\hertz} for six ROIs of 6 pixels each).
    } 
    \afterfigure
    \label{fig:system}
\end{figure}
}

\newcommand{\figTransformer}{
\begin{figure*}[t]
    \centering
    \includegraphics[width=1\textwidth]{ICCV2025-Author-Kit-Feb/FiguresOverleaf/transformers_sketch_v1_compressed.pdf}
    \caption{Vibration Transformer network architecture. Our model processes the signals after conversion to the Fourier domain. The signal from each surface point is processed separately by a shared PointTransformer that extracts information from the spectrum of each point. Then, the added $\pnt{}$ tokens produced by each PointTransformer are processed by a second, ShapeTransformer which integrates the information between all the surface points. Finally, the ShapeTransformer output token is passed through two MLPs where one classifies the container type and the other the liquid fill level. Here LPE stands for `learned positional encoding'.}
    \afterfigure
    \label{fig:network_arch}
\end{figure*}
}

\newcommand{\figExpSetup}{
\begin{figure}[t]
    \centering
    \includegraphics[width=\columnwidth]{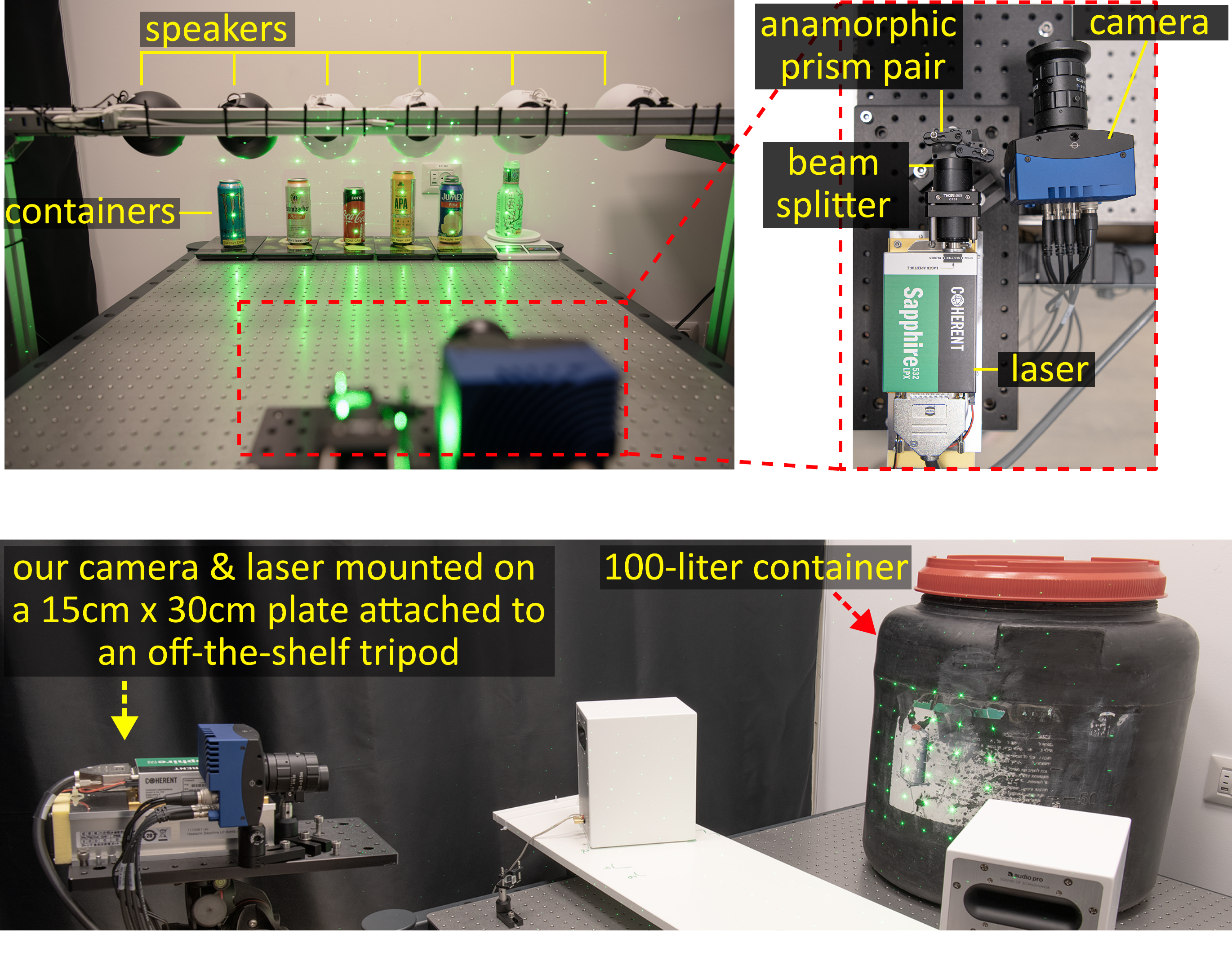}
    \put(-220,89.5){\capfont \textbf{(a)} capturing six containers}
    \put(-80,89.5){\capfont \textbf{(b)} camera system}
    \put(-172,0){\capfont \textbf{(c)} large container experiment}
    \caption{Experimental setup. \textbf{(a)} We acquire vibrations for multiple containers, at multiple points per container at once, for various fill levels. The containers are excited by six different speakers positioned on a beam separated from the container set to avoid transferring mechanical vibrations. Our setup allows simulating sound coming from various directions and amplitudes. We use scales to measure the added liquid level. \textbf{(b)} Vibration sensing system.
    \new{
    \textbf{(c)} We additionally test our system on a large industrial container weighing about \SI{100}{\kilo \gram} when full.
    }
    }
    \afterfigure
    \label{fig:exp_setup}
\end{figure}
}

\newcommand{\figOddcases}{
\begin{figure}[t]
    \centering
    \includegraphics[width=0.95\columnwidth]{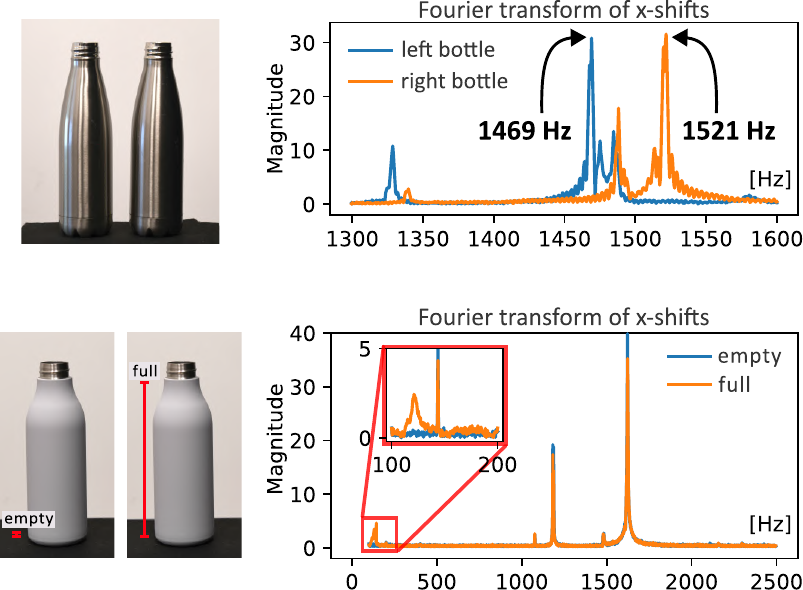}
    \put(-180,87.0){\capfont \textbf{(a)} two \textit{empty} identical silver bottles}
    \put(-205,-10.0){\capfont \textbf{(b)} the same vacuum flask having \textit{different} fill levels}
    \caption{Challenging inference examples.
    \textbf{(a)} Two visually similar \emph{empty} containers that exhibit different resonant frequency profiles, suggesting subtle manufacturing differences. 
    \textbf{(b)} A different container shows nearly identical frequency responses across fill levels, indicating minimal resonance change due to thick insulation. However, noticeable differences do appear in the 100–150~Hz range, enabling inference.
    }
    \afterfigure
    \label{fig:oddcases}
\end{figure}
}

\newcommand{\figDimReduction}{
\begin{figure}[t]
    \centering
    \includegraphics[clip, trim=0mm 24mm 7mm 20mm, width=\columnwidth]{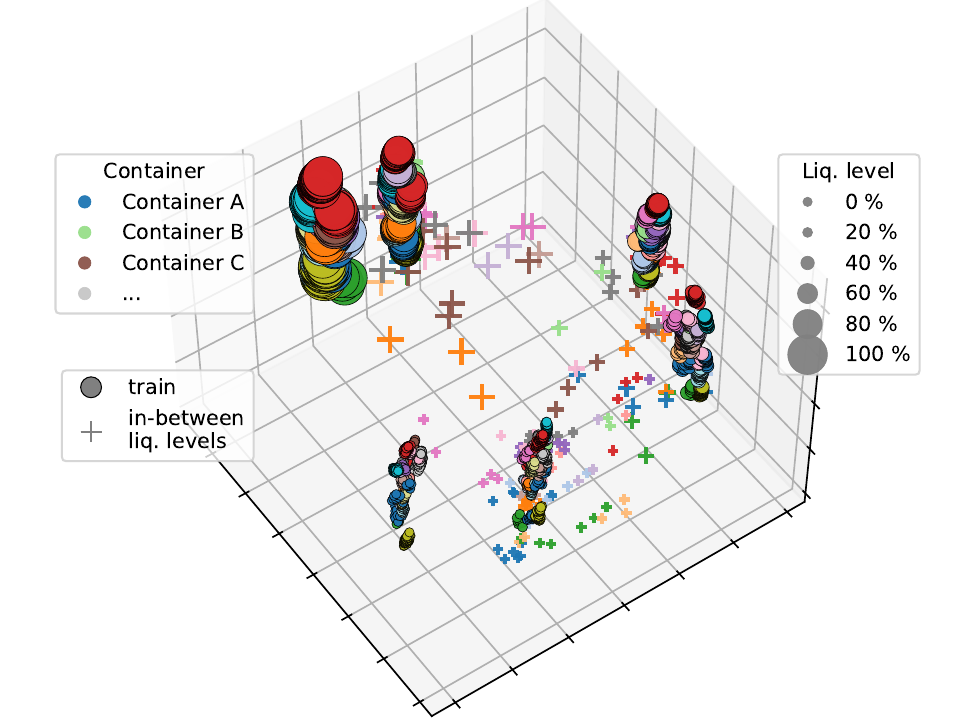}
    \caption{We explore the learned latent space by projecting \cls{} token representations using PCA. Marker size reflects liquid level; color indicates container type. Solid markers (training samples) form distinct fill-level clusters with smooth container-type transitions along the z-axis. Faded markers (``unseen liq. levels" test set) appear interpolated between clusters.}
    \label{fig:pca}
    \afterfigure
\end{figure}
}



%% file: ALL_TABLES.tex
\usepackage{graphicx} 

\newcommand{\blres}[1]{\color{gray}{#1}}  
\newcommand{\err}[1]{{\scriptsize\(\pm\!#1\)}}

\newcommand{\MainResultTable}{
\begin{table}[h]
    \centering
    \small
    \begin{tabular}{l|cc|c}
    \hline
        & \multicolumn{2}{c|}{Level Pred.} & Container \\
        Test name       & Acc. \(\uparrow\) & MAE \(\downarrow\)       & Acc. \(\uparrow\) \\
    \hline
        (a)~within distribution & 0.98 & 0.01 \err{0.03} & 1.00 \\
        ~~~~\blres{CNN baseline} & \blres{0.17} & \blres{0.33 \err{0.22}} &  \blres{0.86} \\
    \hline
        (b)~unseen instances & 0.79 & 0.09 \err{0.19} & 0.95 \\
        (c)~unseen liq. levels & N/A & 0.12 \err{0.11} & 0.81 \\
        (d)~ambient sound      & 0.92 & 0.04 \err{0.12} & 0.97 \\
    \hline
        (e)~unseen liq. levels & \multirow{2}{*}{N/A}       & \multirow{2}{*}{0.15 \err{0.15}}     & \multirow{2}{*}{0.67}   \\
        ~~~~+ ambient sound & & & \\
        (f)~unseen instances       & \multirow{2}{*}{0.59}       & \multirow{2}{*}{0.16 \err{0.23}}     & \multirow{2}{*}{0.77}  \\
        ~~~~+ ambient sound & & & \\
    \hline
    \end{tabular}
    \caption{Experimental validation results. 
    MAE denotes the mean absolute error.
    Liquid levels in our settings are bounded between 0 (empty) and 1.0 (full). 
    Thus, an MAE of \eg~\new{0.01} is equivalent to \new{1}\% error (chance \(\approx\!30\%\)). 
    Container classification accuracy is computed over \new{23} classes, while level accuracy treats level prediction as classification into the predefined set \(L^\text{standard}\)  (not applicable to unseen levels). Test scenarios are presented in order of increasing difficulty, where each subsequent test after (a) requires the model to 
    generalize to additional factors not present in the training set. See the supplementary for detailed results.}
    \label{tab:main_results}
    \vspace{-10pt}
\end{table}
}

\newcommand{\AblationThinResultTable}{
\begin{table}
    \centering
    \small
    \addtolength{\tabcolsep}{0em}
    \begin{tabular}{l|c|c}
    \hline
    Test name & Full model & Single point \\
    \hline
    (a)~within distribution & 0.02 \err{0.05} &  0.03 \err{0.07} \\
    (b)~unseen instances & 0.09 \err{0.15} &  0.11 \err{0.19} \\
    (f)~unseen instances & \multirow{2}{*}{0.16 \err{0.21}} & \multirow{2}{*}{0.18 \err{0.21}} \\
    ~~~~+ ambient sound & &  \\
    \hline
    \end{tabular}    
    \caption{Ablation comparing the full model to a variant that uses only a single surface point measurement.}    
    \label{tab:ablations_thin}
\end{table}
}



%% file: sec/math_macros.tex
\newcommand{\mgrid}{M} 
\newcommand{\ROIh}{P} 

\newcommand{\pnt}{\textsf{[pnt]}\xspace{}}  
\newcommand{\cls}{\textsf{[cls]}\xspace{}}  
\newcommand{\vib}{\bm v}  
\newcommand{\oneShift}{v}  
\newcommand{\Vib}{V}  

\newcommand{\freq}{f}  

\newcommand{\Ntsamples}{N}  

\newcommand{\FFT}[1]{\mathcal{F}\!\left\{ #1 \right\}}

\newcommand{\idxvec}{h}  

%% file: sec/abstract.tex
\begin{abstract}
Computer vision seeks to infer a wide range of information about objects and events. However, vision systems based on conventional imaging are limited to extracting information only from the visible surfaces of scene objects. For instance, a vision system can detect and identify a Coke can in the scene, but it cannot determine whether the can is full or empty.
In this paper, we aim to expand the scope of computer vision to include the novel task of inferring the hidden liquid levels of opaque containers by sensing the tiny vibrations on their surfaces. 
\new{Our method provides a \textit{first-of-a-kind} way to inspect the fill level of multiple sealed containers remotely, at once, without needing physical manipulation and manual weighing.}
First, we propose a novel speckle-based vibration sensing system for simultaneously capturing scene vibrations on a 2D grid of points. We use our system to efficiently and remotely capture a dataset of vibration responses for a \new{variety} of everyday liquid containers.
Then, we develop a transformer-based approach for analyzing the captured vibrations and classifying the container type and its hidden liquid level at the time of measurement. Our architecture is invariant to the vibration source, yielding correct liquid level estimates for controlled and ambient scene sound sources. Moreover, our model generalizes to unseen container instances within known classes (\eg, training on five Coke cans of a six-pack, testing on a sixth) and fluid levels. We demonstrate our method by recovering liquid levels from various everyday containers.
\end{abstract}

%% file: sec/intro.tex
\figTeaser
Since its inception in the 1960s, computer vision has sought to enable machines to infer a wide range of information about scene objects and events. \new{In industrial settings, this capability enables the automation of various inspection and monitoring tasks involving manufactured goods, mechanical parts, or stored items across a range of environments.}
Today, much like in its early days, most computer vision systems rely on sensing using conventional cameras designed to replicate the human eye. However, using conventional cameras limits the scope of retrievable scene information. For instance, an image or video can be used to detect and identify a scene object, but the object's internal properties, like its content or material composition, are mostly indeterminable from its captured surface appearance. In this paper, we focus on revealing one commonplace hidden object property: the fill level of opaque liquid containers.

Our work joins a rich body of previous vision research focused on retrieving hidden object properties by leveraging \textit{unconventional} imaging. 
Clever hyperspectral scene probing and polarization cues were used to classify \cite{saragadam2019krism,lee2024spectral,saragadam2020programmable,saragadam2021sassi} and segment \cite{liang2022multimodal,tominaga2008metal} scene object materials.
Material properties were also probed by thermal imaging \cite{saponaro2015material,cho2018deep,dashpute2023thermal,narayanan2024shape,ramanagopal2024theory}
and structured light methods \cite{liu2013discriminative}. While effective, these methods can only probe object properties that are optically accessible at the object's surface.

To probe deeper, one must use a signal that permeates the object's interior but can be optically sensed at the object's surface. One such signal is object vibrations. In a series of seminal works, Bounman, Davis \etal, captured the minute vibrations of simple objects like rods and fabrics using high-speed cameras to extract properties like object density and Young’s modulus \cite{davis2015visual,bouman2013estimating}. Later, Feng \etal, extended the vibrometry-based method to infer volumetric stiffness and density \cite{feng2022visual,ogren2025visual}. However, capturing high-frequency, low-amplitude vibrations on general object surfaces using high-speed cameras is challenging due to bandwidth and optical magnification constraints. To address these challenges, Sheinin \etal, recently demonstrated modal analysis using a dual-shutter camera prototype that leverages laser speckle to capture vibrations at high speeds for 
colinear scene points \cite{Sheinin:2022:Vibration}. Later, Zhang \etal, used the same system to recover the anisotropy of different-material planar objects \cite{Zhang:2023}.

The visual vibration works described above focused on recovering low-level object properties like motion spectra, material stiffness, or density. In this paper, we seek 
to infer a higher-level object semantic property: the amount of liquid it 
presently holds.
Specifically, similar to Davis \etal~\cite{davis2015visual}, we excite the scene using a nearby speaker and measure the resulting object vibrations using a novel speckle-based imaging system. Our system is inspired by prior speckle-based vibration works \cite{Sheinin:2022:Vibration,zalevsky2009simultaneous,bianchi2019long,wu202120k,wu2020fast,cai2025event2audio}. However, unlike previous works, it can capture a 2D grid of scene points simultaneously, enabling vibration measurement of multiple objects at once, with each object probed at multiple points on its surface (see Figs.~\ref{fig:opening} and \ref{fig:system}).
\new{Notably, several prior works tackled the liquid level recovery task by either recording the sound of liquid pouring \cite{bagad2025soundwaterinferringphysical,Wilson2019}, or recording the sound resulting from a physical knock on the container \cite{garcia2018non}. Conversely, our approach eliminates the need for any physical interaction with the container or reliance on nearby microphones, enabling passive, remote inference of liquid content using only visual measurements.}

Inferring the hidden liquid level from a container's vibrations is a challenging task.
The vibrational response of an object having a simple shape and material composition can be modeled using a small set of material parameters (\eg, mass and stiffness). As shown in prior works, these parameters can be recovered by observing the object's resonant frequencies and mode shapes \cite{davis2015visual,feng2022visual}. While some liquid containers, such as certain wine glasses, exhibit a simple relationship between the fluid level and the resulting resonant frequencies, most everyday containers feature complex geometries and heterogeneous materials, resulting in a non-trivial relationship between the liquid level and the resulting vibrations. 
The challenge becomes substantial when inferring fluid levels of unseen containers from the same class, as slight manufacturing variations can shift resonant frequencies even among identical-looking items (\eg, cans in a six-pack).
To address this challenge, we develop a learning-based approach introducing a novel physics-inspired `Vibration Transformer' to classify the fluid level of various everyday liquid containers. The Vibration Transformer receives a spectral decomposition of the recorded two-axis, multi-point surface vibrations. It is, thus, invariant to the speaker excitation (content and duration).\footnote{As long as the excitation signal is sufficiently broadband.}

\new{Our approach introduces a novel way to remotely assess the fill levels of multiple sealed containers at once without requiring any physical handling or manual weighing.
We believe our method could facilitate the inspection of large warehouses storing consumer liquid products (\eg, soda cans, shampoos), heavy containers impractical to weigh, and containers holding hazardous, toxic, flammable, or radioactive liquids (\eg, heavy or tritiated water). 
The latter, long-storage containers, are especially susceptible to spillage and evaporation over time—issues that are further exacerbated by repeated physical handling and inspection.}

To evaluate our approach, we gather a dataset of various everyday liquid containers and measure their vibration response at three surface points per container for multiple speaker positions and short (two-second) excitations (\eg, a logarithmic chirp, a segment of a popular song, and ambient noises). 
We show that our approach can accurately classify the hidden liquid levels of the dataset containers for novel speaker positions and fluid levels not seen during training. Moreover, we show that if trained on multiple same-class containers (\eg, five Coke cans), the model can infer the hidden liquid levels of containers outside the training set (\ie, a sixth Coke can).


%% file: sec/back.tex
\subsection{Modeling object vibrations}
\label{sec:vibration_model}
Under small deformations, most objects can be approximated as linear elastic, meaning they exhibit a proportional relationship between stress and strain and return to their original shape once the disturbance is removed. The vibrations of objects in this regime can be modeled using a second-order linear differential equation called the \textit{equation of motion}, which depends on the object's material properties (\ie, mass, stiffness, and damping), boundary conditions, and external forces (\ie, the vibration excitation) \cite{den1985mechanical, chen2015modal}.

In certain conditions, the vibrations of linear elastic objects can be expressed as a linear combination of their \textit{resonant frequencies}, or vibration modes \cite{den1985mechanical}.
Measuring these modes—defined by a set of independent natural frequencies and mode shapes—enables inference of the equation of motion parameters mentioned above \cite{davis2015visual,feng2022visual}.
For everyday liquid containers, the equation of motion must account for the coupled interaction between the container and the liquid inside. The presence of liquid introduces added mass, increasing inertia and lowering resonant frequencies (see Fig.~\ref{fig:wine}). Additionally, fluid-structure interactions alter stiffness and damping, resulting in complex vibrations with a non-trivial relationship to the container's liquid level. Furthermore, slight manufacturing differences in the same container type may yield different resonance frequencies even for the same fluid level (Fig.~\ref{fig:oddcases}(a)). Therefore, in this work, we use a learning-based approach to infer this complex relation, for various container types and instances.
\figWine

\subsection{Measuring surface vibrations}
\label{sec:speckle}
Surface vibrations can be measured using various principles, including high-speed imaging \cite{chen2015modal,davis2015visual,feng2022visual}, interferometry (\eg, Laser Doppler Vibrometers (LDV)) \cite{rothberg1989laser}, contact-based piezoelectric transducers \cite{lichtenwalner1998local}, and more. In this paper, we rely on speckle-based vibrometry, which consists of illuminating an object's surface points with a laser and capturing a defocused video of the illuminated point \cite{zalevsky2009simultaneous,smith2017colux}. 

Due to the laser light coherency, the defocused laser spot will contain a random interference pattern called \textit{speckle}~\cite{zalevsky2009simultaneous, alterman2021imaging}. This speckle pattern is highly sensitive to small surface tilts, causing it to shift within the point's defocused spot \cite{zalevsky2009simultaneous,Sheinin:2022:Vibration}. Thus, unlike LDVs, speckle-based vibrometry obviates the need for expensive specialized hardware and simplifies vibration sensing to computing image-domain shifts. In Sec.~\ref{sec:camera}, we describe a novel vibration-sensing method allowing vibration capture for a 2D grid of surface points.

%% file: sec/cam.tex
Our vibration sensing system measures vibrations for a 2D grid of scene surface points. 
As such, it enables simultaneous multi-point vibration sensing for multiple scene containers. For simplicity, Sec.~\ref{sec:camera} and \ref{sec:learning} focus on a single container, though the same method is independently applied to each scene container.
Our system operates on a simple yet effective principle, capturing speckle vibrations robustly at high speeds in previously undemonstrated configurations.

\figSystem

As shown in Fig.~\ref{fig:system}(a), our prototype consists of a single laser and a camera. The laser illuminates the scene through a custom diffractive beam splitter that splits the beam into a 2D grid (\eg, $6 \times 6$). The split beam then passes through an anamorphic prism pair, configured to widen the beam angles along the horizontal axis, thereby aligning the laser points with the scene's container arrangement (see Fig.~\ref{fig:exp_setup}).

The camera is defocused to yield a grid of speckle patches (Fig.~\ref{fig:system}(b)). Recovering the two-axis vibrations at each sensed laser grid point involves computing the image-domain shifts of the speckle within each point's patch~\cite{zalevsky2009simultaneous}. However, at full resolution, the camera's bandwidth severely limits the maximum frame rate, resulting in insufficiently fast vibration sampling rates. Therefore, to achieve high sampling rates, we configure the camera's readout to output only $\mgrid$ regions of interest (ROIs) of size $ W \! \times\! \ROIh$ pixels, which speeds up the camera's FPS by approximately a factor of $H/(\mgrid\ROIh)$, where $H$ and $W$ are the camera's full image resolution height and width, respectively. For example, the camera in Fig.~\ref{fig:system} can operate at \SI{2247}{\hertz} at full resolution. However, for $\mgrid\!=\!6$ ROIs of height $\ROIh\!=\!6$ pixels, the camera FPS jumps to \SI{57699}{\hertz}, a rate sufficient for most mechanical and acoustic vibration applications. See Sec.~\ref{sec:details} for full hardware details and  Sec.~\ref{sec:discussion} for a discussion relating our system to prior works. 

To robustly compute the two-axis image-domain shifts, \( \oneShift\!\in\!\mathbb{R}^2 \), between every two consecutive frames, we develop an ad-hoc method that first uses phase-correlation (PC) \cite{Kuglin:1975:PhaseCorr,Pearson:1977:videoPhaseCorr} to recover integer-pixel shifts, followed by a Lukas-Kanade (LK) \cite{Lucas:1981:LK} estimation of the residual sub-pixel translation (similar to \cite{Douini:2017:PCLK}). 
To handle the high volume of shift computations for recovering multiple laser points at high frame rates (\eg, 1.44 million calls for \SI{2}{\sec} at \SI{20}{\kilo\hertz} with a \(6 \times 6\) grid), we implement a parallelized batched GPU version of PCLK, called PCLK+, to streamline vibration recovery.
Our GPU implementation is \(\times 20\)~faster.

Let $\vib_i\!\in\! \mathbb{R}^{2\times \Ntsamples}$ denote the two-axis image-domain speckle shifts recovered by our camera at surface point $i$, where $i\!\in\!\{0,1,2\}$ and $\Ntsamples$ is the number of time samples at camera sampling rate $f_{\rm cam}$ [Hz].\footnote{WLOG, we assume each container is illuminated with 3 laser points.}
The measured signal $\vib_i$ has units of pixels and can be converted to surface tilts by multiplying with a per-axis scalar \cite{Sheinin:2022:Vibration}. However, since our method does not require such calibration, for simplicity, we will refer to $\vib_i$ as the object \textit{vibrations} at point~$i$.
Next, we describe how to extract the container's fill level given the set of measured container vibrations $\{\vib_0,\vib_1,\vib_2\}$.

%% file: sec/learning.tex

\figTransformer
We aim to recover the container’s liquid level from raw vibration signals \(\{\vib_0,\vib_1,\vib_2\}\) recorded at three surface points.
Here, we describe the Vibration Transformer~-- a transformer-based model that takes the multi-point vibration signals \( \vib_i \) and outputs the liquid level as a percentage of the container's total capacity (\eg, \SI{20}{\percent} full).

While prevailing signal processing frameworks operate in the temporal (\eg~\cite{van:2016:WaveNet,ravanelli:2018:sincnet}) or the short time Fourier transform (STFT) domains (\eg~\cite{Ephrat:2018:lookingToListen,Amodei:2016:deepspeech,chan:2015:listen_attent_spell,Hershey:2017:audioClass}), motivated by classic modal analysis, where an object's vibrational modes are recovered to infer low-level object material characteristics
, we operate solely in the Fourier domain.
That is, the input to our model is 
\begin{equation}
 \Vib_i\left[ \freq \right] \equiv |\FFT{\vib_i}|  \in \mathbb{R}^2, \freq \in F^{\rm fixed}
 \label{eq:fft}
\end{equation}
where $\FFT{}$ denotes the Discrete Fourier Transform, $\freq$ is frequency, and $F^{\rm fixed}$ is a predefined frequency set (\eg, \mbox{$F^{\rm fixed} = \{40,41,42,..,2500\}$} [Hz]). Note that the definition of Eq.~\eqref{eq:fft} allows for $\vib_i$ of an arbitrary duration, content and sampling frequency $f_{\rm samp} > 2\max(F^{\rm fixed})$.

The complexity of regressing the multi-point multi-axis Fourier response $ \Vib_i\left[ \freq \right],i \in \{0,1,2\}$, to the container's liquid level can vary greatly between different container types. In rare cases, a container's Fourier analysis can easily relate to the liquid level. For example, some wine glasses have a highly resonant response, yielding a distinct high-frequency audible tone that predictably decreases in frequency as the glass is filled (see Fig.~\ref{fig:wine}(a)). However, none of the containers in our dataset have such simple characteristics. Rather, we found that most everyday containers exhibit complex spectral responses that are difficult to interpret and regress using classical methods (\eg, Fig.~\ref{fig:wine}(b)). Therefore, we base our model on the popular Transformer architecture \cite{Vaswani:2017:Attention}, letting the model learn the non-trivial relation between the vibrations at multiple points and the liquid level.

Further inspired by modal analysis, our Vibration Transformer comprises two main components. In the first stage, a shared \textit{PointTransformer} independently processes the signal from each point, analyzing the frequency responses and local resonance characteristics (\ie, mode frequencies). Then, a subsequent {\textit{ShapeTransformer} fuses the information from these local features, enabling the model to reason about the mode shapes. 
Ultimately, the Vibration Transformer's latent representation captures both liquid level and container type from the raw vibration input.

\parvspace\paragraph{Vibration Transformer architecture.}


A schematic of our model is shown in Fig.~\ref{fig:network_arch}.
The frequency representation for each point \( \Vib_i \)  is divided into fixed-sized non-overlapping tokens via a Tokenizer module, where each token encodes information about a specific frequency band $\Delta f$.
A learnable positional encoding is added to each token, and the sequence of input frequency tokens is supplemented with a trainable \pnt{} token \cite{Devlin:2019:BERT}.
The PointTransformer uses self-attention to extract information from all frequency bands into the \pnt{} token per point $i$, independently.

Then, the ShapeTransformer processes the sequence of three `transformed' tokens  \( \pnt{}_i \). Here, additional positional encoding is added to each token \( \pnt{}_i \), encoding the grid position of point \( i \) on the container.
A trainable \cls{} token is added to integrate the information from all measured points.
We use two MLPs, one to infer the container type, represented as a discrete label \( c \), and another for
the discrete liquid level $l$. In this work, we set \new{\mbox{\(l \!\in\! L = \left\{0.0, 0.2, 0.4, 0.6, 0.8, 1.0\right\} \).}}

We train our model using supervised learning. 
Each collected training sample includes \( \left\{ \vib_0\!\left(t\right), \vib_1\!\left(t\right),\vib_2\!\left(t\right); c, l \right\} \). 
Our network produces two probability distribution vectors: one for the container class \(\hat{c}\!\in\!\mathbb{R}^{K_{\rm cont}}\) and one for the liquid fill level $\hat{p}\!\in\!\mathbb{R}^{6}$.
Let $\idxvec\!=\!1,2,..,6$ denote the index within vector $\hat{p}$ and the corresponding index within the ordered set~$L$.
In contrast to the container class prediction, which we optimize using a standard cross-entropy loss, the inherently ordinal nature of the liquid fill level leads us to employ a variation of the Sorted ORDinals (SORD) Loss~\cite{Diaz:2019:sord0,Roy:2020:sord1,Frank:2021:sord2}.
For a given true fill level \( l \in L \), we define a soft target distribution vector $q_l\in\mathbb{R}^{6}$, where
\begin{equation}
q_l[\idxvec] = \left(e^{-50\left(l-L[\idxvec]\right)^2}\right)/\sum_j e^{-50\left(l-L[j]\right)^2},
\end{equation}
The SORD Loss is then given by:
\begin{equation}
\mathcal{L}_{\text{SORD}} = - q_l^T \cdot \log \left( \hat{p} \right).    
\end{equation}
The overall loss is a weighted sum of the cross-entropy loss for \( c \) and the SORD Loss for \( l \).

We use the output logits vector \(\hat{p} \) to predict the liquid level in two ways. We use a Maximum a posteriori (MAP) estimator to predict the most likely discrete liquid level 
\begin{equation}
\hat{l}_\text{MAP} \equiv L[h_\text{MAP}], ~~h_\text{MAP} = \arg\max_\idxvec \hat{p}[\idxvec].
\end{equation}
Additionally, we wish to test the model's behavior for liquid levels in between the levels of $L$. In such cases, we compute the expectation over the prediction 
\begin{equation}
    \hat{l}_\mathbb{E} = \sum_{\idxvec}L[\idxvec] \cdot \hat{p}[\idxvec]
    \label{eq:cont}
\end{equation}
Note that the model is trained to predict the discrete levels \(\hat{l}_\text{MAP}\), while \( \hat{l}_\mathbb{E}\) can be used to output values $\in \left[0, 1\right]$.

%% file: sec/details.tex
\subsection{Hardware details}
Our system comprises an EoSens2.0MCX12 camera \cite{eosens2.0mcx12-fm} and a Coherent Sapphire \SI{532}{\nano\meter} \SI{500}{\milli\watt} laser \cite{sapphire-lpx-532}. The laser is passed through a HOLO-OR beamsplitter, yielding a $6 \!\times\! 6$ point grid with a separation angle of $2.75 \!\times\! 2.75$ degrees \cite{holo-or-beamsplitter}. Thus, at \SI{500}{\milli\watt}, each point has a \SI{13.9}{\milli \watt} power. To illuminate a row of six containers at once, we pass the laser grid through an unpaired Thorlabs anamorphic prism pair \cite{thorlabs-anamorphic-prism}. The pair is adjusted to create the desired horizontal spread, turning the square laser grid into a rectangular one.

To probe the row of containers, we create an array of six Creative Pebble V2 speakers \cite{pebble-v2-speakers-pair}, which can be individually activated via a ten-channel MOTU UltraLite-mk5 audio interface \cite{motu-sound-card}. The speakers are mounted on a frame detached from the containers, which sit on a vibration-isolated optical breadboard, thus minimizing the transition of mechanical vibrations through the table (see Fig.~\ref{fig:exp_setup}(a)).

\subsection{Training the Vibration Transformer}
%
First, we compute the DFT magnitude, per axis, for each vibration signal \( \vib_i \) on the predefined frequency set 
\(
F^\text{fixed} = \{ \SI{100}{}, \SI{100.5}{}, \ldots, \SI{2500}{\Hz} \}
\).
The resulting \( \Vib_i \) is a \(2 \times 4800\) matrix of Fourier magnitude coefficients. We then divide this matrix into non-overlapping patches of size \(2 \times 100\) (\ie, each patch contains 200 coefficients). For each patch, we apply a linear projection to map the 200-dimensional coefficient vector into a 512-dimensional token. This produces a sequence of 48 tokens, which are subsequently fed into our PointTransformer.
%
%
We add learnable position encoding to the tokens.
%
Our PointTransformer and ShapeTransformer have eight transformer layers each, with four heads in the self-attention layers. 
There are about 25 million trainable parameters in each transformer.
The prediction MLPs have one hidden layer with 64 dimensions and ReLU activation.

%
We use the Adam~\cite{Kingma:2014:adam_opt} optimizer with a learning rate \(\lambda\!\! =\!\! 10^{-5}\) to train the model for 7500 epochs.
%
During training, we augment the data by simulating random smooth frequency responses that modulate the magnitudes of the Fourier coefficients (\ie, filters)~-- mimicking variations in speaker type and acoustic environmental characteristics~-- 
and randomly drop \SI{50}{\percent} of the PointTransformer input tokens.
%
We give \(\mathcal{L}_\text{SORD}\) a weight of 0.9 and 0.1 for \(\mathcal{L}_\text{CE}\).

%% file: sec/experiments.tex
\figExpSetup
To gather data for training and testing our model, we built an experimental rig comprised of our camera, a set of six speakers positioned on a beam above the test table, and a set of six kitchen scales to measure the amount of liquid we add to each container (see Fig.~\ref{fig:exp_setup}). In each data sampling iteration, we place a set of six containers on the scales filled to one of the liquid levels in \mbox{\new{\( L^{\rm standard} = \left\{0.0, 0.2,0.4, 0.6, 0.8, 1.0 \right\} \)}}. Then, we continuously capture their vibrations at three surface points per container while playing multiple short audio sequences using one or more of the six speakers. While the camera can capture at much higher speeds, we set its frame rate to $5100$Hz, having observed that most containers have little energy at frequencies higher than $2000$Hz. A subset of containers were captured with extra fill level \mbox{\new{$L^{\rm interm} = \left\{0.25, 0.50, 0.75 \right\}$}}.

\new{In addition to the small-sized everyday containers described above, our dataset also includes one large 100-liter industrial container (see Fig.~\ref{fig:exp_setup}(c)).
This container has a 0.53-meter diameter and weighs about \SI{100}{\kilo \gram} when full, and thus cannot be easily placed on any off-the-shelf scale.}
In total, \new{our dataset contained 5910 individual data samples}. Please see the supplementary materials for a full dataset breakdown and visualization.

We partitioned the container dataset into subsets to test the model's prediction for several increasingly complex inference tasks. We train our model on \new{23} unique container types to predict the levels in $L^{\rm standard}$, using two types of excitations: a two-second logarithmic chirp with start and end frequencies of $100$ and $2500$Hz, respectively, and a two-second segment of a popular song.  
Then, we test our model in six validation categories: (a) \textit{within-distribution}, (b) \textit{unseen instances}, (c) \textit{unseen liquid levels}, (d) \textit{ambient sound}, (e) \textit{unseen levels at ambient sound} and (f) \textit{unseen instances at ambient sound}.
\footnote{
\new{Since the container of Fig.~\ref{fig:exp_setup}(c) is a single unique instance, it was only included in tests \textit{(a)} and \textit{(d)}. See the supplementary for more details.}}

\MainResultTable
\parvspace\paragraph{Test (a): within-distribution.} Here we test the model's ability to \new{predict} the fill level of containers included in the training set,
 but for novel speaker positions never seen by the model during training. This corresponds to a task of measuring the liquid level of non-disposable containers that have been previously `seen' by the system and require repeated testing (\eg, in factories or offices where fill levels must be monitored regularly). To effect this test, we randomly exclude one of the six speakers for each of the training set containers. This designates about 20\% of our data for testing. Tab.~\ref{tab:main_results} shows that our model excels at this task, 
 yielding only \new{1}\% mean absolute error (MAE) on the test set.

\parvspace\paragraph{Test (b): unseen instances.} 
Here we test the model on five new containers, each similar in type to those in the training set but not seen during training (\eg, training on five cans of a six-pack, and testing on the sixth). 
The train set contained three to seven examples for each tested container. Our model achieves good predictions, with 9\% error over the test set. Notably, as shown in Fig.~\ref{fig:oddcases}, we found there could be manufacturing variations between seemingly identical containers, yielding a deviation in the spectral responses. Thus, for category (b), we expect that more training examples should better capture class statistics and improve performance.

\parvspace\paragraph{Test (c): unseen liquid levels.}
Here, we test how well our model generalizes to liquid levels outside the training distribution. Despite only being trained to predict levels in $ L^{\rm standard}$, we hypothesized that the ordinal nature of our prediction would lead to reasonable predictions outside $ L^{\rm standard}$. 
To test this category, we applied the model to containers with $L^{\rm interm}$, predicting the fill level using the $\hat{l}_\mathbb{E}$ estimator (Eq.~\eqref{eq:cont}). Results show \new{12\%} test error.

\parvspace\paragraph{Test (d): ambient sound.}
Our model was trained on containers \emph{actively} excited by a nearby speaker. However, most environments already contain ambient noises.
In this category, we tested the model by playing ambient supermarket background noise. 
Simulating the ambient sounds using speakers was necessary because our lab is quiet by design. 
Evaluation on this unseen, structureless excitation yields good predictions (4\% error), suggesting our Fourier-based approach is largely invariant to the excitation audio.

Lastly, for completion, tests (e) and (f) contain additional combinations of cases (b),(c), and (d) with increasingly complex inference tasks. Nevertheless, our model performs reasonably, even under these edge scenarios (\eg, unseen liquid levels tested using ambient sound). Overall, our model provides good liquid level predictions despite being trained on a relatively small dataset. Therefore, we believe that increasing the dataset by orders of magnitude would improve all the tests in Tab.~\ref{tab:main_results}.

\parvspace\paragraph{CNN-based baseline.}
We tested several naive approaches before adopting the transformer architecture in Fig.~\ref{fig:network_arch}. 
Notably, we implemented an eight-layer convolutional neural network (CNN) applied independently to each \(\vib_i\), followed by eight more layers on the concatenated hidden representations. Each layer uses a kernel size of 15, BatchNorm~\cite{ioffe:2015:batch_norm}, and ReLU activation.
The first three layers in each component also perform stride-2 average pooling with a 3x-sized kernel.
Overall, the CNN had 27M parameters in the first component and 40M in the second, \new{roughly matching the Vibration Transformer's parameter count}.
Tab.~\ref{tab:main_results} shows this approach classifies container types reasonably well but fails to recover liquid levels, performing no better than chance.

\new{
\parvspace\paragraph{Discrete vs.~continuous liquid level prediction.} 
Replacing classification over $L$ levels trained with $\mathcal{L}_{\mathrm{SORD}}$ by a single continuous output trained with $\ell_1$ loss raises the MAE from 0.01 to 0.20 on the within-distribution test.
 }

\parvspace\paragraph{Does phase play a pivotal role?} The input to our network is the signal \emph{magnitude}, discarding phase information (Eq.~(\ref{eq:fft})). 
We explored versions where we retain the phase by either concatenating it to the magnitude or by processing the raw complex $\FFT{\vib_i}$. 
Neither performed better than our base model.
See the supplementary for detailed results.

\AblationThinResultTable
\parvspace\paragraph{How many points?}
Our prototype collects three vibration measurements per container. To evaluate the impact of multi-point data, we trained models using only one measurement while having the same number of transformer layers.
Tab.~\ref{tab:ablations_thin} shows that while a single point suffices in a within-distribution setting, harder cases~-- like unseen instances of known containers~-- can benefit from multi-point data which encapsulates the mode shapes. 

\figDimReduction
\parvspace\paragraph{Exploring the learned latent space.}
Fig.~\ref{fig:pca} visualizes our model's internal representation of the input data. The plot is generated by extracting the \cls{} tokens from all the training samples and 
applying PCA to project these embeddings onto three principal components. The result reveals six distinct, elongated vertical clusters matching the six discrete liquid fill levels present in our training set. Within each cluster, container types are not randomly scattered; instead, they exhibit an organized shift along the z-axis, with a gradual transition from one container type to another. The plot also contains 
the samples from the ``unseen liquid levels” test (displayed as faded markers).
Their position between clusters suggests that, despite training on discrete levels, the model learns a latent representation capturing the continuous spectrum of liquid fill levels.
\new{The clear structure shown in Fig.~\ref{fig:pca} suggests the model learned meaningful features rather than overfitting to individual training examples, despite our dataset's relatively small scale.}

%% file: sec/discussion.tex
\newcommand{\dparvspace}{\vspace*{0pt}}
\figOddcases
\paragraph{Advantages and limitations relative to prior speckle vibrometry systems:}
Our system was inspired by many prior works that reduced the image domain to increase capture speed \cite{antipa2019video,Sheinin:2020:Diffraction,weinberg2020100}.
Closest to our work is the dual-shutter camera of Sheinin \etal, which can capture multiple surface points on a single container. However, since it senses only a single row, capturing multiple containers requires scanning.
Conversely, our system can sense multiple containers at once, making it more practical for applications that require scanning large item sets (\eg, supermarkets, assembly lines). 
Beyond scanning 2D grids, our system obviates the two-camera calibration of~\cite{Sheinin:2022:Vibration}, avoids light loss from beam splitting, and suffers no rolling-shutter inter-frame dead times.
Nevertheless, combining both approaches (\ie, adding a second reference camera) can enable reducing the ROI heights to a single pixel, maximizing camera speed. 

\dparvspace\paragraph{Performance analysis on various container types:}
Our experiments offer insights into the model’s performance across container types.
Firstly, we observed that highly resonant containers are \emph{more} difficult to infer for unseen same-class containers. 
\new{This is because their frequency response may be dominated by a strong resonant peak, whose frequency position can shift due to manufacturing differences (see Fig.~\ref{fig:oddcases}(a)). 
For example, consider a hypothetical limiting case where two same-class containers have distinct resonant frequencies of \SI{450}{\Hz} and \SI{400}{\Hz}, respectively. In this scenario, the model would be unable to distinguish between an empty second bottle and the first containing some liquid, as liquid lowers the resonant frequency. 
Real containers, however, do not exhibit perfect delta-like frequency responses, making them amenable to our approach (as in Fig.~\ref{fig:oddcases}(a)).}
We also observed that some vacuum flasks yield similar vibrations across fill levels due to double-wall insulation (Fig.~\ref{fig:oddcases}(b)). Fortunately, the response differs enough at lower frequencies, allowing for a reasonable inference.

\dparvspace\paragraph{Container materials, laser safety, and audio level:}
We tested our method on everyday containers and captured speckle interference in most cases without modifying the packaging. However, some materials, like glass, polished metals, and ones having a very low albedo, will be less amenable to speckle-based vibrometry. For such materials, like the wine glass in Fig.~\ref{fig:wine}, we placed a small sticker on the container's surface to capture the speckle. Such augmentation can be readily applied for sensing non-disposable containers (\eg, in industrial factory settings).
Each laser point in our prototype had \SI{14}{\milli\watt} power, making direct eye exposure unsafe.
\footnote{Diffuse reflections at \SI{14}{\milli \watt} are generally considered safe.} 
Thus, eye safety must be considered due to potential direct exposure from specular surfaces.
\new{
We evaluated how sound volume affects vibration signal SNR (see supplementary). Results show good SNR at low volumes, suggesting the method works with minimal sound sources (\eg, small compact speakers).}

\dparvspace\paragraph{Generalizing to novel container classes:}
We present a proof-of-concept for generalizing liquid inference to unseen speaker positions, fluid levels, input sounds, and containers of the same class. However, our dataset is too small to explore broader questions: Can a model trained on enough data generalize to entirely new container classes? Classify liquid types (\eg, water, soda, oil) or extend to granular materials (\eg, sand)? Can vibrations serve as a `container fingerprint' for identifying the same container across scenes?  
We leave these fascinating questions for future work.

%% file: sec/conclude.tex
We introduced a novel system that ``sees" inside opaque liquid containers by combining a novel imaging system based on high-speed laser speckle vibrometry with a new deep learning architecture~-- the Vibration Transformer~-- for semantic analysis of vibration signals. 
We conducted an extensive experimental evaluation and provided insightful analysis of the results to validate our approach and demonstrate a proof of concept for this novel computer vision task. Our work also yielded a novel container vibration dataset that may be beneficial to the vision community beyond the scope of the current work. 
We hope our work inspires further research on the semantic inference of hidden scene properties, such as detecting the contents of closed packages, fruit ripeness, spoilage in sealed foods, chemical composition, and other latent attributes.